\documentclass[10pt]{article}
\usepackage{microtype}
\usepackage{graphicx}
\usepackage{subfigure}
\usepackage{booktabs} 
\usepackage{amsmath}
\usepackage{gensymb}
\usepackage{longtable}
\usepackage{caption}
\usepackage{graphicx}
\usepackage{multicol}
\usepackage{comment}
\usepackage{natbib}
\usepackage[english]{babel}
\PassOptionsToPackage{hyphens}{url}\usepackage{hyperref}
\usepackage{float}
\usepackage{hyperref}

\usepackage{siunitx}
\usepackage{etoolbox}
\newcommand{\PLH}{{\mkern-2mu\times\mkern-2mu}}
\newcommand{\tinypm}{\raisebox{1.0pt}{$_{^\pm}$}}
\newcommand{\ubold}{\fontseries{b}\selectfont}
\robustify\ubold

\newcommand\blfootnote[1]{%
  \begingroup
  \renewcommand\thefootnote{}\footnote{#1}%
  \addtocounter{footnote}{-1}%
  \endgroup
}
\usepackage[accepted]{icml2020}

\usepackage[accepted]{icml2020}
\def\doubleunderline#1{\underline{\underline{#1}}}
\icmltitlerunning{Uncertainty in Neural Relational Inference Trajectory Reconstruction}

\begin{document}

\twocolumn[
\icmltitle{Uncertainty in Neural Relational Inference Trajectory Reconstruction}




\begin{icmlauthorlist}
\icmlauthor{Vasileios Karavias}{phys}
\icmlauthor{Ben Day}{compsci}
\icmlauthor{Pietro Li\`{o}}{compsci}
\end{icmlauthorlist}

\icmlaffiliation{phys}{Department of Physics, University of Cambridge.}
\icmlaffiliation{compsci}{Department of Computer Science and Technology, University of Cambridge}

\icmlcorrespondingauthor{Ben Day}{ben.day@cl.cam.ac.uk}

\icmlkeywords{Machine Learning, ICML}

\vskip 0.3in
]



\printAffiliationsAndNotice{} 

\begin{abstract}
Neural networks used for multi-interaction trajectory reconstruction lack the ability to estimate the uncertainty in their outputs, which would be useful to better analyse and understand the systems they model. In this paper we extend the Factorised Neural Relational Inference model to output both a mean and a standard deviation for each component of the phase space vector, which together with an appropriate loss function, can account for uncertainty. A variety of loss functions are investigated including ideas from convexification and a Bayesian treatment of the problem. We show that the physical meaning of the variables is important when considering the uncertainty and demonstrate the existence of pathological local minima that are difficult to avoid during training.
\end{abstract}

\section{Introduction and Related Work}
\label{submission}
Deep learning models have been found to work well at tasks such as classification \citep{Alex,Nam} and data processing \citep{Kumar}. Recently, Neural Networks (NNs) have been found to perform well at retrieving physical laws \citep{cranmer} and in particular in many-body multi-interaction trajectory reconstruction tasks where the model also infers the relations between particles \citep{kipf,webb}. Yet these implementations do not estimate errors in their outputs, which are essential to analysing the inferred interaction models, calibrating equipment, and allow for better informed engineering \& better physical understanding of the system.\par
In this paper, the factorised Neural Relational Inference (fNRI) \citep{webb} is extended to output both a mean and standard deviation for the position and velocity predictions which, together with an appropriate treatment by the loss function, can account for uncertainty. The methods presented can trivially be applied to the NRI model as well.\par
\begin{figure*}[h]
     \centering
    \includegraphics[width=200pt]{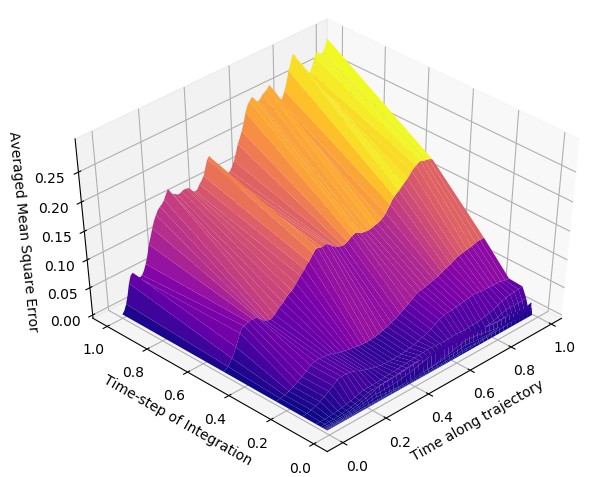}
     \hfill
    \includegraphics[width=200pt]{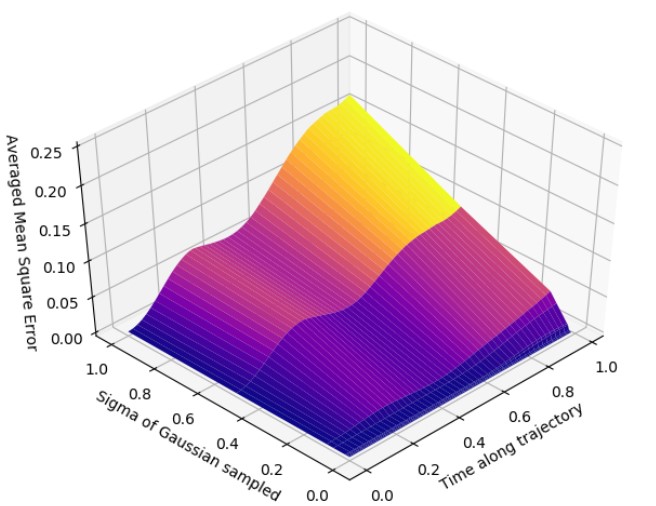}
     \hfill
      \caption{\textbf{Left}: Computational errors over time and time-step of integration. The general trend is that the errors increase with increasing time along the trajectory and time-step of integration. The dips are likely due to random fluctuations and a remnant of the leapfrog method used for integrating Newton's equations. \textbf{Right}: Physical errors over time and standard deviation of the Gaussian used to generate initial deviations from the ground truth. The general trend is that the errors increase with increasing time along the trajectory and $\sigma$ of the Gaussian used to obtain the perturbations in initial conditions. Curiously, both forms of error are of the same order of magnitude and thus neither can be ignored in the analysis. It should be noted that the same arbitrary units are used in both plots.}
    \label{fig:two_graphs}
\end{figure*}

There are three main expected sources of error. First are the \textbf{computational errors} which include rounding errors from handling floating point numbers and which propagate throughout the model, which are typically small, and a larger contribution arising from the model having a finite time-step.
Errors from the use of finite time-step integration methods increase with increasing time-step and this will factor into the model \citep{riley}. The second form of errors are \textbf{physical errors}. These are explained by chaos theory as a consequence of non-linear interactions such as the Coulomb force\citep{CH:1}, which states that for small changes in initial conditions, particle deviation grows exponentially \citep{AL1}. The expected value of both errors can be investigated using the simulator and the results are shown in Figure \ref{fig:two_graphs}. The final error source is the \textbf{model error} which is difficult to estimate. This is an open problem and the main topic of this investigation. \par
Recent papers have investigated teasing out model uncertainty by using dropout in the network as a Bayesian deep Gaussian process \citep{Gal,kendall}. Models have been developed to introduce uncertainty in the weights of the NN to deal with noisy data \citep{Blundell,pawlowski}. There have also been investigations into predicting and reducing errors in the output with various NN structures \citep{Nima}.\blfootnote{The full code along with a readme file to reproduce the experiments presented can be found at \url{https://github.com/vassilis-karavias/fNRIsigma-master}}\par


\section{Methods}
The fNRI \citep{webb} builds on the neural relational inference (NRI) model from \citet{kipf}. The NRI is formulated as a generalisation of the variational autoencoder (VAE).
In the fNRI, a simulator generates trajectory data by integrating Newton's laws for randomly sampled interaction networks to generate train, validation and test sets. The model takes the trajectory positions and velocities of the particles and uses the encoder to output an edge type posterior distribution from which the edge types can be obtained as samples of the distribution. The decoder takes the initial positions and the interaction graph and reconstructs the trajectories by outputting a likelihood distribution of the particle positions and velocities.
This allows the output variations to give a measure of the uncertainty but do not actually output an uncertainty value. Training is carried out by minimising the loss $$L = \sum_{i} \left(\frac{(\vec{\hat{y}}_{i}-\vec{y}_{i})^{2}}{2\sigma^{2}}\right) - D_{kl}(q_{\theta}(\vec{z}|\vec{y})||p(\vec{z})) = L_y+L_{KD}$$
where the sum is over the time-steps, particles and batches, and $\vec{\hat{y}}$ represents the predicted positions and velocities while $\vec{y}$ represents the true values. The second term is the Kullback-Leibler divergence. Note that in \citet{webb} the uncertainty $\sigma$ is fixed at $\sigma^{2}= 5 \times 10^{-5}$. For more details of the NRI model see \textbf{figure 2}.
\begin{figure*}[h]
    \centering
    \includegraphics[trim = 0 105 0 0, clip, width=\textwidth]{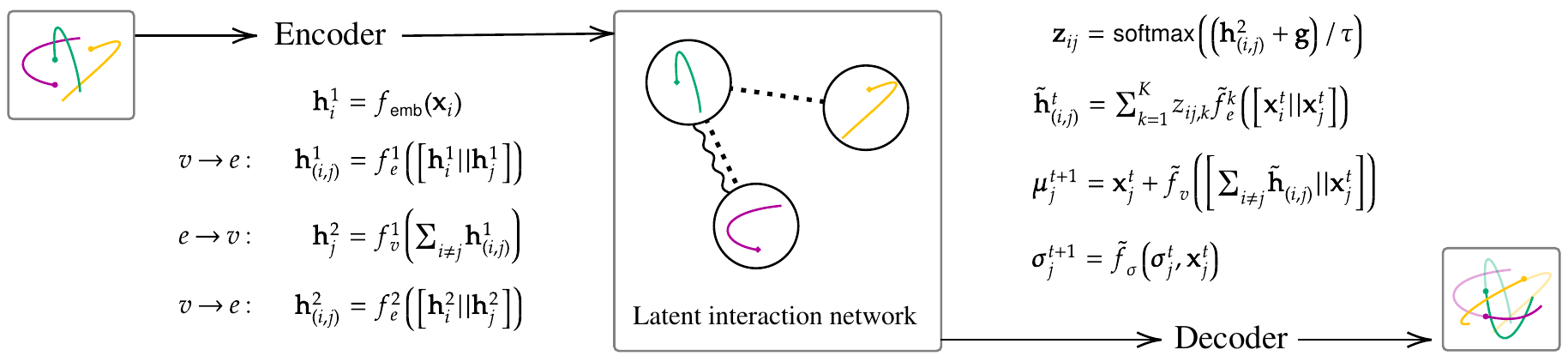}
    \caption{NRI-type model overview \citep{kipf}. The encoder embeds trajectories ($\mathbf{x}$) and, using vertex-to-edge ($v \rightarrow e$) and edge-to-vertex ($e \rightarrow v$) message-passing operations, produces the latent interaction network. The sampled edges ($\mathbf{z}$) modulate pairwise functions ($\tilde{f}^k_e$) in the decoder that can be associated with forces in classical physics. A function of the net resultant `force' (sum over $k$) is used to update the mean position using a skip-connection. $[x||y]$ indicates concatenation.}
    \label{fig:model_overview}
    \vspace{-15pt}
\end{figure*}
\subsection{Directly Predicting Uncertainty}
The data is based on spring and charge interactions between 5 particles in a finite 2 dimensional box, as in \citep{webb}, and as such $\vec{y}$ has four coordinates $(x,y,v_{x}, v_{y})$. We allow the value of $\sigma$ in the loss function to vary. There are 3 cases of interest: 
\begin{itemize}
    \setlength\itemsep{0.0em}
    \item isotropic in all four coordinates;
    \item semi-isotropic, that is, isotropic in position ($x,y$) and velocity ($v_x,v_y$) separately;
     \item anisotropic in all four coordinates.
\end{itemize}
The isotropic model was not expected to perform well as it involves fitting a single $\sigma$ to four different parameters with two different physical dimensions. This could work if the main error source is from the model, and thus the physical semantics of the parameters can be ignored.
The first term in the loss function for the anisotropic case is modified to
\begin{align}
L_y=\frac{1}{2}\sum_{i}\left(\ln{||\doubleunderline{\Sigma}_{i}||} + (\vec{\hat{y}}_{i}-\vec{y}_{i})^{T}\doubleunderline{\Sigma}_{i}^{-1}(\vec{\hat{y}}_{i}-\vec{y}_{i})\right)
\end{align}
where $\doubleunderline{\Sigma}$ is the covariance matrix. In the isotropic case this reduces to
\begin{align}
L_y= \sum_{i}\left(\frac{(\vec{\hat{y}}_{i}-\vec{y}_{i})^{2}}{2\sigma_{i}^{2}} + \frac{1}{2}\ln{\sigma_{i}^{2}}\right)
\end{align}
The variation in $\sigma$ is predicted by adding it as a parameter to the NN phase space vector.
There has been work on convexifying the loss landscape, which allows for quicker convergence to the minimum \citep{paquette}. We show that convexification can be used to improve the performance of the models tested.\par

\subsection{Bayesian Approaches}
Assuming the output of the model follows a Gaussian distribution with mean $\vec{y}_{i}$ and variance $\doubleunderline{\Sigma}_{i}$ we can take a Bayesian approach to model fitting. This takes the form of an additional Kullback-Leibler term between the prior and the output distribution. The expected distribution depends on the time-step and is centred about the true position, i.e. $\vec{\mu}_{i} = \vec{y}_{i}$. The $\sigma$ of the distribution is assumed to be $\sigma_{i} = \sqrt{\Delta x_{0}^{2}+\Delta x_{comp,t}^2 + \Delta x_{phys,t}^2}$.
$\Delta x_0$ is the average deviation of positions and velocities between time-steps 0 and 1. The other terms represent the computational and physical errors discussed in section 2 and were calculated using those methods. $\Delta x_{comp,t}$ was determined using the error data for the time-step of the model. $\Delta x_{phys,t}$ depends on a preexisting error and the computational error is used as a seed to calculate the expected physical error.
The Normal-Inverse-Wishart distribution was also tested. This is the conjugate prior to data sampled from a multivariate Normal distribution with unknown mean and covariance matrix \citep{conjNIW}. This is exactly the case here. The posterior to the distribution is also a Normal-Inverse-Wishart distribution with scaled hyper-parameters \citep{conjNIW}. The loss function was taken as the negative logarithm of the posterior.
\subsection{Z-score}
To obtain a quantitative analysis of how good a measure of the error the $\sigma$ values obtained are, the z-score of each data point was calculated. The z-score is defined as: $\vec{z}_{i} = \frac{\vec{\hat{y}}_{i} - \vec{y}_{i}}{\sigma_{i}}$.
 The z-score distribution gives a measure of the quality of the uncertainty estimate. A good estimate will have a unit Normal z-score distribution, provided the underlying distribution is Gaussian.
 
\section{Results}

\begin{table*}[h]
\small
    \centering
        \caption{\footnotesize Accuracy ($\%$) in inferring interaction graph. The higher the better. MSE/ $10^{-5}$ in trajectory reconstruction for the different models. The smaller the better. The fixed variance model can be seen to do the best when considering trajectory reconstruction and does only slightly worse than the Lorentzian Loss model with $\sigma^2_0 = 5\times 10^{-5}$ in inferring an interaction graph. Overall, this model does better than others for these conditions. However, it cannot predict the uncertainty in the output. The z-score distribution was fit with a Gaussian and Lorentzian and a quality of fit test was performed. The best fit has been displayed along with the value of the test. The smaller the value of the quality of fit the better the fit is. The models that do not have a fit are because the distributions are not appropriate for a fit to neither Lorentzian nor Gaussian. It is particularly interesting that most z-score distributions are fit better to a Lorentzian rather than a Gaussian. The quality of fit tends to vary.}
        \vskip 0.15in
        \label{tab:edge_acc}
        \sisetup{%
            table-align-uncertainty=true,
            separate-uncertainty=true,
            detect-weight=true,
            detect-inline-weight=math
        }
        \begin{tabular}{l|c*{2}{S[table-format=2.3]@{\,\(\tinypm \)\,}S[table-format=2.3]}|c*{1}{S[table-format=2.3]@{\,\(\tinypm \)\,}S[table-format=1.4]}}
            \toprule
            Model & $\sigma_0^2$ & \multicolumn{2}{c}{Accuracy $/ \%$} & \multicolumn{2}{c|}{\textsc{mse} $/ \ 10^{-5}$} & Best fit to Z-score & \multicolumn{2}{c}{Quality of Fit} \\
            \midrule
            fNRI baseline & $5 \PLH 10^{-5}$ & 91.506 & 0.002 & 6.08 & 0.03 & - & \multicolumn{2}{c}{-}\\
            \midrule
            Isotropic Gaussian & $5 \PLH 10^{-5}$ & 29.6 & 1.4 & 64.24 & 0.05 & Lorentzian & 0.93 & 0.02 \\
            Isotropic Gaussian: Convex. & $5 \PLH 10^{-5}$ & 86.29 & 0.02 & 8.3 & 0.2 & Lorentzian & 11.9 & 0.2 \\
            Isotropic Gaussian & $1 \PLH 10^{-10}$ & 89 & 3 & 7.4 & 0.4 & - & \multicolumn{2}{c}{-}\\
            Isotropic Lorentzian & $5 \PLH 10^{-5}$ & 93.2 & 0.2 & 9.2 & 1.1 & Gaussian & 0.499 & 0.004 \\
        \vspace{1mm}
            Isotropic Lorentzian & $1 \PLH 10^{-10}$ & 53 & 3 & 130 & 20 & Lorentzian & 1.136 & 0.002 \\
        \vspace{1mm}
            Semi-isotropic Gaussian & $5 \PLH 10^{-5}$ & 84.7 & 1.0 & 8.3 & 0.3 & Lorentzian & 16.96 & 0.09 \\
            Anisotropic Gaussian & $5 \PLH 10^{-5}$ & 86 & 2 & 102.1 & 0.5 & Lorentzian & 15.8 & 1.4 \\
            Anisotropic Gaussian & $1 \PLH 10^{-10}$ & 84.3 & 0.2 & 7.1 & 0.2 & - & \multicolumn{2}{c}{-} \\
            Anisotropic Gaussian-KL & $5 \PLH 10^{-5}$ & 53.57 & 0.06 & 58.2 & 0.2 & Lorentzian & 0.611 & 0.004 \\
            Anisotropic Gaussian-KL & $1 \PLH 10^{-10}$ & 87 & 2 & 7.8 & 0.5 & Lorentzian & 0.390 & 0.002 \\
            \bottomrule
        \end{tabular}
    
\end{table*}

\textbf{Table 1} presents the performance metrics of the various models tested. The fixed variance fNRI model \citep{webb} is given as a benchmark.\par
\subsection{Gaussian Models}
 The isotropic gaussian model with  $\sigma_0^2 = 5 \times 10^{-5}$ struggles to learn the interaction graph and only manages to recreate the trajectory for the first few time-steps. The z-score fit shows a decent fit to a Lorentzian distribution rather than a Gaussian but the errors were overestimated. Adding a convexifying term to the loss function yields much better interaction graph inference and trajectory reconstruction and thus improves the general performance of the model. However, the z-score results showed that the model could not predict the uncertainties and were not particularly well fit by either distribution. With $\sigma_0^2 = 1 \times 10^{-10}$, the interaction graph inference and trajectory reconstruction are much better. This is contrary to our expectation that the model would do badly. However, this can be explained as this model learned to keep $\sigma$ constant and thus it does not fit a $\sigma$ value to four coordinates. This also means the performance metrics are similar to the benchmark. Keeping $\sigma$ constant is undesirable for predicting uncertainties.\par

The isotropic Gaussian model displays difficulties in reconstructing trajectories and inferring an interaction graph with  $\sigma_0^2 = 5 \times 10^{-5}$. This could be due to the model trying to fit a single $\sigma$ to too many parameters as explained earlier. The semi-isotropic model performs much better than the isotropic Gaussian with the same initial $\sigma_0$, suggesting that the physical semantics of the parameters in question cannot be ignored when modelling the uncertainty in these physical systems. The z-score distribution fits better to a Lorentzian than a Gaussian but the fit is not particularly good. In fact, this model also keeps $\sigma$ approximately constant.\par
The anisotropic case with $\sigma_0^2 = 5 \times 10^{-5}$ shows good interaction graph inference however trajectories are not successfully reconstructed, with the model only being able to predict the first few time-steps effectively. The z-score distribution showed the model was overestimating the uncertainty which in turn allowed the model to keep the trajectory far apart without introducing a large contribution to the loss. Starting with $\sigma_0^2 = 1 \times 10^{-10}$ fixes the trajectory reconstruction issue, but mostly because the model learns to keep $\sigma$ fixed, and the uncertainty is therefore not predicted.
\subsection{Lorentzian Loss}
The Lorentzian distribution was empirically found to be a better fit to the z-score distribution than the Gaussian distribution as shown in table 1. It was then considered that perhaps the underlying distribution was Lorentzian and not Gaussian.  The $\sigma$ output of the NN was redefined as the full-width half-maximum (FWHM), $\gamma$, of the Lorentzian distribution and the loss function became $$L_y=\sum_{i}(\ln{(1+\frac{(\vec{\hat{y}}_{i}-\vec{y}_{i})^{2}}{\sigma_{i}^{2}})} + \ln{\sigma_{i}})$$
Results show that the $\sigma_0^2 = 5 \times 10^{-5}$ model successfully infers an interaction graph and reconstructs the trajectories. However, the z-score distribution is skewed, meaning there is some bias to the uncertainty predictions, and therefore the predictions are not indicative of the true uncertainty. Despite this, this distribution is fit well to a Gaussian distribution and is one of the more successful models at predicting the uncertainty. This bias is removed when using $\sigma_0^2 = 1 \times 10^{-10}$ but the interaction graph inference and trajectory reconstruction performance is much worse.\par
\subsection{Bayesian Models}
The addition of the Kullback-Leibler term described in section 2.2 to the anisotropic Gaussian allows us to add prior knowledge of the errors obtained from the simulator shown in Figure \ref{fig:two_graphs}. With an initial $\sigma_0^2=5\times 10^{-5}$, the model cannot learn to reconstruct trajectories and struggles to infer the interaction graph. The uncertainty is not predicted well either, displaying a clear bias as in the Lorentzian loss case. However, switching to an initial $\sigma_0^2 = 1 \times 10^{-10}$ yields much more reasonable performance but learns to keep $\sigma$ fixed despite the decent fit to a Lorentzian. The Normal-Inverse-Wishart distribution was also considered for the reasons discussed in section 2.2. However, the model failed to converge in any training example and thus was empirically found to be a bad model to use.

\section{Final comments}
Results show that pathological models are difficult to avoid since the models fall into local minima. The specific pathological model found depends on the loss function used and the initial parameters, but the models typically fell into one of 2 local minima: 
\begin{itemize}
    \setlength\itemsep{0.0em}
    \item the model learns to keep a fixed variance and infer the interaction graph and reconstructs the trajectory well
    \item the model over-estimates $\sigma$ and does not reconstruct trajectories well after a certain time
\end{itemize}
It was shown that convexification can be used to avoid certain bad pathological models, although the model fell into a different type of pathology by reaching a different local minimum, albeit a more desirable one.\par
Further investigation is required to understand how the initial parameters effect the training of the model. Also, the effects of convexification could be investigated further \citep{paquette}. Alternatively, the fNRI model could be trained with fixed $\sigma$, then feed the resulting data into a new NN to obtain the uncertainty. More appropriate priors also need to be investigated to see their effects on the resulting model.
\clearpage

\newpage
\twocolumn[
    \begin{@twocolumnfalse}
    \icmltitle{Uncertainty in Neural Relational Inference Trajectory Reconstruction: \\
    Supplementary Material}
    \end{@twocolumnfalse}
]

\section*{Overview}
These supplementary materials are provided to support the Graph Representation Learning and Beyond (GRL+) NeurIPS 2020 workshop paper `Uncertainty in Neural Relational Inference Trajectory Reconstruction'.
The material provides further details of the experimental setup.

\section*{Experimental Setup}
The 2-layer multi-layer perceptron (MLP) fNRI model developed in \citet{webb} was used, along with the aforementioned extensions.\par

\subsection*{Implementation}
The Neural Networks presented in this work were implemented in Python using PyTorch \citep{paszke}. The implementation is based on the fNRI model developed in \citet{webb} which in turn was developed based on the implementation of the NRI model developed in \citet{kipf}. Over 2000 lines of original code was written to add the functionality needed for this paper. The full source code for the paper can be found at \url{https://github.com/vassilis-karavias/fNRIsigma-master} and contains a readme file that explains how to recreate the experiments in this paper.
\subsection*{Simulations}
The majority of the details are the same as those used in \citet{webb}, those that are different are explicitly stated as such.\par
We use N=5 point particles in a finite size 2D box, with elastic wall collisions and no external background potential. It was found that the model without a box did not train well. This was likely due to the normalisation process using the largest values in each of the coordinates. If the particles are not contained in the box it is likely that there are cases where the particle coordinates become much larger than what is typically observed and as such the typical coordinates will have `normalised' values that are orders of magnitudes smaller. This will be detrimental to training, and thus the box helps prevent this problem. The initial positions are sampled from a Gaussian distribution $N(0,0.5)$ and the initial velocity was a randomly chosen vector with a norm of $0.5$. Particles interact through spring and charge forces, with pairs of particles connected by springs at random (Bernoulli sampling over edges with $p=0.5$), and particles either charged or not at random (Bernoulli sampling over particles with $p=0.5$).
(This corresponds to the I+C case in \citet{webb}.) Newton's equations are solved using a leapfrog method with a time-step of $0.001$ sampled at a frequency of $100$ so the model gets data sampled at a time-step of $0.1$. $50,000$ training simulations, $10,000$ validation simulations and $10,000$ test simulations were generated.

\subsection*{Model}
We used the Adam optimiser algorithm \citep{kingma} with an initial learning rate of 0.0005, which halved every 200 epochs. The models were trained for 500 epochs with a batch size of 128. The value of the loss functions defined by each of the model was used as a measure of when to checkpoint. A softmax temperautre of $\tau =0.5$ was used in the concrete distribution \citep{maddison}.\par
The hidden and output dimensions used were 32, which is different to the 256 used in  \citet{webb} as that was found to use too much memory, the model used batch-normalisation and ELU activation functions. The encoder trained on the first 50 time-steps and the decoder trained on the next 50 time-steps. During training of the decoder the ground truth state is fed back to the decoder every 10 time-steps. The fixed variance model used a variance of $5\times 10^{-5}$.
\subsection*{Normalisation of $\sigma$}
It should be noted that $\sigma$ must be positive and, since the data is normalised, it is preferable to transform $\sigma$ to a value in the range [-1,1] and re-transform back using a smooth function with positive values. This results in better training and performance. This was done by passing the initial $\sigma_{0}$ through an inverse softplus function $x = \frac{1}{\beta}\ln{|e^{\beta \sigma_{0}}-1|}$. $x$ was then passed into the NN and the output, $y$, was passed through a softplus $\sigma = \frac{1}{\beta}\ln{(1+e^{\beta y})}$ to get the output $\sigma$. Two different $\sigma_0$ values were investigated: $\sigma_0^2 = 5\times10^{-5}$ with $\beta = 5$ and $\sigma_0^2 = 1\times10^{-10}$ with $\beta = 10$.
\clearpage

\section*{Acknowledgements}
 The authors would like to thank Edoardo Calvello for his explanation of convexification and design of an algorithm that can be used to test this in future. The authors wish to thank Ezra Webb and Helena Andres-Terre for making the codebase for the fNRI model \citep{webb} publicly available as well as Thomas Kipf, Ethan Fetaya, Kuan-Chieh Wang, Max Welling and Richard Zemel for making the NRI model \citep{kipf} codebase publicly available. This research was only possible thanks to their commitment to open research practices. Finally the authors wish to thank the developers of PyTorch \citep{paszke} and \url{https://www.mathcha.io/}.

\bibliography{references.bib}
\bibliographystyle{icml2020}

\end{document}